\documentclass{article}

\usepackage[preprint]{neurips_2026}

\usepackage[utf8]{inputenc} 
\usepackage[T1]{fontenc}    
\usepackage{hyperref}       
\usepackage{url}            
\usepackage{booktabs}       
\usepackage{amsfonts}       
\usepackage{nicefrac}       
\usepackage{microtype}      
\usepackage{xcolor}         
\usepackage{hwemoji}

\title{How Open Must Language Models be to Enable Reliable Scientific Inference?}

\usepackage{cleveref}
\usepackage{booktabs}
\usepackage{multirow}
\usepackage{float}

\newcommand\OURopenweight{\textsc{Open-Weight Model}}
\newcommand\OURminmodel{\textsc{Closed Model}}

\newcommand\OURopenweights{\textsc{Open-Weight Models}}
\newcommand\OURminmodels{\textsc{Closed Models}}

\newcommand\OUReval{\textsc{Model Evaluation}}
\newcommand\OURcomp{\textsc{Model Comparison}}
\newcommand\OURinterp{\textsc{Model Interpretability}}

\usepackage{tikz}
\usetikzlibrary{calc}
\usepackage{xcolor}

\definecolor{cYes}{HTML}{4CAF50}
\definecolor{cPrinciple}{HTML}{FF9800}
\definecolor{cSometimes}{HTML}{FFC107}
\definecolor{cRarely}{HTML}{FFC107}
\definecolor{cNo}{HTML}{F44336}
\definecolor{hdrBg}{HTML}{37474F}
\definecolor{rowA}{HTML}{F5F5F5}
\definecolor{rowB}{HTML}{FFFFFF}
\definecolor{sepLine}{HTML}{B0BEC5}
\definecolor{dividerLine}{HTML}{546E7A}

\author{
\textbf{James A. Michaelov\textsuperscript{1}}, 
\textbf{Catherine Arnett\textsuperscript{2}}, 
\textbf{Tyler A. Chang\textsuperscript{3}}, \\
\textbf{Pamela D. Rivi\`{e}re\textsuperscript{4}}, 
\textbf{Samuel M. Taylor\textsuperscript{3}}, 
\textbf{Cameron R. Jones\textsuperscript{5}}, 
\textbf{Sean Trott\textsuperscript{4}}, \\
\textbf{Roger P. Levy\textsuperscript{1}}, 
\textbf{Benjamin K. Bergen\textsuperscript{3}}, 
\textbf{Micah Altman\textsuperscript{1}} \\
\textsuperscript{1} Massachusetts Institute of Technology, 
\textsuperscript{2} EleutherAI, 
\textsuperscript{3} University of California San Diego,\\
\textsuperscript{4} Rutgers University-Newark,
\textsuperscript{5} Stony Brook University
}

\begin{document}

\maketitle

\begin{abstract}
When pretrained language models are used in scientific research, the choice of which to use is often left uninterrogated. However, in many cases, the extent to which valid scientific inferences can be drawn from a given experiment hinges on assumptions that can only be made on the basis of information that is only currently available for open-weight models. In this paper, we provide a framework for considering how suitable a closed or open-weight model is for a given research question, advice on how to resolve or mitigate the issues incurred by the use of closed models, and a checklist to aid in reporting, reviewing, and careful reading with respect to these key methodological details.
\end{abstract}

\section{Introduction}
\label{sec:introduction}

In the last decade, language models have shifted from being primarily scientific artifacts such as BERT \citep{devlin_2019_BERTPretrainingDeepb} and T5 \citep{raffel_2020_ExploringLimitsTransfera} to being incorporated into products such as ChatGPT and Gemini that are used by hundreds of millions of people \citep{cai_2025_AlphabetPlansMassive,kant_2025_OpenAIsWeeklyActive,musumeci_2025_GooglesGeminiUsage}. Given that they are competing products, developers of such systems report ever-increasingly high performance at a range of tasks, while information about these systems has become increasingly scarce as part of preserving competitive advantages. Many have raised concerns about the lack of transparency around such models, often from the perspectives of accountability, governance, safety, reliability, and hindrances to innovation (see, e.g., \citealp{mitchell_2019_ModelCardsModel,bommasani_2023_FoundationModelTransparencyc,longpre_2024_LargescaleAuditDataset}). However, given the reported state-of-the-art performance of these commercial products, it is unsurprising that they have made their way into scientific research in a wide range of fields \citep[see, e.g.,][]{pramanick_2026_TransformingScholarlyLandscapes}.

But the increasingly blurring line between scientific artifact and commercial product comes with risks for scientific research. Specifically, the fact that very little information is known about the design and training of widely-used commercial systems such (Chat)GPT, Gemini, and Claude, and the fact that they are ever-evolving products that are optimized based on user feedback and their wider reception has led a number of researchers to question whether there is a place for such commercial systems in research at all \citep{liesenfeld_2023_OpeningChatGPTTracking,rogers_2023_ClosedAIModels,palmer_2024_UsingProprietaryLanguage}. 

However, there is little scientific guidance for determining when it is appropriate to use a closed model in research. To the best of our knowledge, no previous work has directly analyzed when it is a problem to use closed models. Furthermore, with the exception of ACL Rolling Review's guidance that reviewers do not ask for closed models as baselines, we are not aware of any policies on the study of closed language models at research venues. Finally, recent surveys have demonstrated that the use of closed models in research is common, and in many cases, growing (\citealp{balloccu_2024_LeakCheatRepeat,kalyan_2024_SurveyGPT3Family,chen_2026_LLMassistedSystematicReview}; see also \citealp{movva_2024_TopicsAuthorsInstitutions,ye_2025_LLMs4AllReviewLarge,pramanick_2026_TransformingScholarlyLandscapes})

The main contribution of our paper is to close this gap: we provide a theoretical framework for considering how the use of closed models threatens the validity of a range of types of scientific claim, systematically analyze the extent to which open weights and other possible interventions can resolve and mitigate these issues, and provide specific guidance and recommendations for researchers, readers, and reviewers (see \Cref{fig:summary_table}). Fundamentally, we argue that given the importance of these questions, the choice of whether to use a more open or closed language model in a given study is a critical methodological decision and should be treated as such. To aid in the adoption of our recommendations we provide a sample checklist in Appendix \ref{app:checklist}.

\begin{figure}
    \centering
\begin{tikzpicture}[
    every node/.style={inner sep=0pt, outer sep=0pt},
    font=\sffamily\footnotesize,
]

\newcommand{\colAw}{4.6}
\newcommand{\colBw}{4.7}
\newcommand{\colCw}{4.7}
\newcommand{\rowH}{1.03}

\pgfmathsetmacro{\xB}{\colAw}
\pgfmathsetmacro{\xC}{\colAw+\colBw}
\pgfmathsetmacro{\TW}{\colAw+\colBw+\colCw}

\newcommand{\suitcell}[7]{%
  \fill[#5, rounded corners=2pt]
    ({#1+0.06}, {#2+0.05}) rectangle ++({#3-0.12}, {#4-0.10});
  \pgfmathsetmacro{\txtW}{#3-0.3}%
  \node[text=white, align=center, text width=\txtW cm,
        inner sep=0pt]
    at ({#1+#3/2}, {#2+#4/2}) {%
      {\sffamily\footnotesize\bfseries #6}\\[0pt]
      {\sffamily\scriptsize\baselineskip=7.2pt\relax #7\par}};
}

\newcommand{\plaincell}[6]{%
  \pgfmathsetmacro{\txtW}{#3-0.2}%
  \node[align=center, text width=\txtW cm, inner sep=0pt,
        font=\sffamily\footnotesize\bfseries]
    at ({#1+#3/2}, {#2+#4/2}) {#5\\[-1.5pt]#6};
}

\pgfmathsetmacro{\hdrH}{\rowH}
\pgfmathsetmacro{\hdrY}{6*\rowH}
\fill[hdrBg, rounded corners=3pt]
  (0, \hdrY) rectangle ++(\TW, \hdrH);

\pgfmathsetmacro{\hdrMid}{\hdrY+\hdrH/2}
\pgfmathsetmacro{\txtWA}{\colAw-0.3}
\pgfmathsetmacro{\txtWB}{\colBw-0.3}
\pgfmathsetmacro{\txtWC}{\colCw-0.3}

\node[text=white, font=\sffamily\footnotesize\bfseries,
      align=center, text width=\txtWA cm]
  at ({\colAw/2}, \hdrMid) {Inferential Goal};
\node[text=white, font=\sffamily\footnotesize\bfseries,
      align=center, text width=\txtWB cm]
  at ({\xB+\colBw/2}, \hdrMid) {Is a Closed Model\\[-1pt]Sufficient?};
\node[text=white, font=\sffamily\footnotesize\bfseries,
      align=center, text width=\txtWC cm]
  at ({\xC+\colCw/2}, \hdrMid) {Is an Open-Weight\\[-1pt]Model Sufficient?};

\foreach \i/\bg in {
  5/rowA, 4/rowB,
  3/rowA, 2/rowB, 1/rowA, 0/rowB}{
  \fill[\bg] (0, {\i*\rowH}) rectangle ++(\TW, \rowH);
}

\foreach \i in {1,...,5}{
  \draw[sepLine, thin] (0, {\i*\rowH}) -- ++(\TW, 0);
}
\foreach \x in {\xB, \xC}{
  \draw[sepLine, thin] (\x, 0) -- ++(0, {6*\rowH});
}
\draw[hdrBg, line width=0.8pt, rounded corners=2pt]
  (0,0) rectangle (\TW, {6*\rowH});

\plaincell{0}{5*\rowH}{\colAw}{\rowH}
  {Existence Proof}{}
\suitcell{\xB}{5*\rowH}{\colBw}{\rowH}{cYes}
  {Yes}{}
\suitcell{\xC}{5*\rowH}{\colCw}{\rowH}{cYes}
  {Yes}{}

\plaincell{0}{4*\rowH}{\colAw}{\rowH}
  {Solution Candidate}{Generation (Verifiable)}
\suitcell{\xB}{4*\rowH}{\colBw}{\rowH}{cYes}
  {Yes}{}
\suitcell{\xC}{4*\rowH}{\colCw}{\rowH}{cYes}
  {Yes}{}

\plaincell{0}{3*\rowH}{\colAw}{\rowH}
  {Model Evaluation}{}
\suitcell{\xB}{3*\rowH}{\colBw}{\rowH}{cPrinciple}
  {Only in principle}{Requires mitigating the versioning problem}
\suitcell{\xC}{3*\rowH}{\colCw}{\rowH}{cYes}
  {Yes}{}

\plaincell{0}{2*\rowH}{\colAw}{\rowH}
  {Model Comparison}{}
\suitcell{\xB}{2*\rowH}{\colBw}{\rowH}{cPrinciple}
  {Only in principle}{Requires mitigating versioning \& credit assignment problems}
\suitcell{\xC}{2*\rowH}{\colCw}{\rowH}{cYes}
  {Yes}{}

\plaincell{0}{1*\rowH}{\colAw}{\rowH}
  {Model Interpretability,}{Behavioral}
\suitcell{\xB}{1*\rowH}{\colBw}{\rowH}{cPrinciple}
  {Only in principle}{Requires mitigating versioning, credit assignment, \& info restriction problems}
\suitcell{\xC}{1*\rowH}{\colCw}{\rowH}{cYes}
  {Yes}{}

\plaincell{0}{0*\rowH}{\colAw}{\rowH}
  {Model Interpretability,}{Mechanistic}
\suitcell{\xB}{0*\rowH}{\colBw}{\rowH}{cNo}
  {No}{}
\suitcell{\xC}{0*\rowH}{\colCw}{\rowH}{cSometimes}
  {Sometimes}{May require additional information,\\ e.g., data or training code}

\end{tikzpicture}
    \caption{A summary of our main conclusions. \textsc{Model Evaluation}, \textsc{Model Comparison}, and \textsc{Model Interpretability} are defined in \S\ref{ssec:use_cases}; the Versioning Problem (\S\ref{ssec:versioning}), Credit Assignment Problem (\S\ref{ssec:credit_assignment}), and Information Restriction Problem (\S\ref{ssec:output_only}) are defined in their respective sections.}
    \label{fig:summary_table}
\end{figure}

\subsection{Defining \OURopenweights{}}
\label{ssec:openness}
Researchers have proposed a wide range of standards for defining the extent to which a model is ``open'' or ``closed'' \citep{solaiman_2023_GradientGenerativeAI,bommasani_2023_FoundationModelTransparencyc,liesenfeld_2024_RethinkingOpenSource,white_2024_ModelOpennessFrameworka,opensourceinitiative_2024_OpenSourceAI,basdevant_2024_FrameworkOpennessFoundation}. These standards are designed to support a diverse operational, economic, legal, social, and inferential goals, and as such, their specific features vary substantially. We primarily focus on the distinction between closed models and those for which the weights are provided.

\paragraph{Closed Models} Information is most restricted in what we define as a \OURminmodel{}: a system that accepts input texts supplied through a human-computer interface (e.g., a `chat' interface); processes these based on a hidden procedure that may involve pre-processing, one or more language models, post-processing, and possibly additional components (see \S\ref{ssec:credit_assignment}); and emits output text. This definition of a \OURminmodel{} corresponds to the `Hosted Access' category of models identified by \citet{solaiman_2023_GradientGenerativeAI}, and may also be referred to as a fully `black-box' model. An example of such a system is ChatGPT as accessed through a browser. Although model providers often supply additional information about the underlying model(s), we argue that in practice, such information does not generally resolve the key issues that \OURminmodels{} present for reliable inference.

\paragraph{Open-Weight Models} \OURminmodels{} contrast with \OURopenweights{}, which we define as models for which the complete weights and tokenizer are provided, along with a full accounting of the language model architecture and sufficient code to run the model on arbitrary input---i.e., to output the full probability distribution that the model calculates based on a given user-defined input. This aligns with the `downloadable weights' category of models identified by \citet{solaiman_2023_GradientGenerativeAI}.

\subsection{Language models in scientific research}
\label{ssec:use_cases}

We consider how restrictions on the information provided for \OURminmodels{} limit the inferences that can be made when carrying out scientific research using such models. Crucially, our focus is on research where the goal is to make inferences about models that generalize to future behavior to some degree. This can range from the expectation that a future execution of the same model on similar data should be comparable in some way to broader general claims about language models. We thus do not consider issues related to the use of language models in paper preparation \citep{kusumegi_2025_ScientificProductionEra}. While our analysis can be applied to a wider range of scientific research contexts, we specifically focus on three inferential goals that underlie a substantial amount of research involving language models, as detailed below.

\paragraph{(i) Model Evaluation} We use the term \OUReval{} broadly to encompass all empirical tests of what a model \textit{does}; i.e., how one or more models behave given a specific input. This includes evaluating the performance of a given language model on specific benchmarks \citep[e.g.][]{wang_2019_GLUEMultiTaskBenchmarka,liang_2023_HolisticEvaluationLanguage,srivastava_2023_ImitationGameQuantifyinga,biderman_2024_LessonsTrenchesReproducibleb}, analyzing patterns of model behavior that may or may not be explicitly part of the task \citep[e.g.][]{rahmati_2025_CoCoCoLaEvaluatingImprovinga}, cases where there is no ground truth \citep[e.g.][]{basar_2025_TurBLiMPTurkishBenchmarka,gonen_2025_DoesLikingYellowa}, or other similar analyses of model outputs.

\paragraph{(ii) Model Comparison} We use the term \OURcomp{} to refer to any study carrying out \OUReval{} on two or more models and comparing their performance, including similarities and differences between them. Model comparison is thus dependent on \OUReval{}---reliable comparisons require reliable evaluations.

\paragraph{(iii) Model Interpretability} We use the term \OURinterp{} to refer to research with the goal of understanding the mechanisms underlying language model behavior.
In this category, we include studies that aim to determine whether specific behaviors (e.g., better performance) are attributable to specific design features such as architectural features or training data \citep[e.g.,][]{penedo_2024_FineWebDatasetsDecantinga,yang_2024_ParallelizingLinearTransformersa,qiu_2025_GatedAttentionLarge,yue_2025_DoesReinforcementLearning}. We also include research on scaling laws and work aimed at characterizing higher-level `capabilities' that underlie other behaviors \citep[e.g.,][]{kaplan_2020_ScalingLawsNeuralb,wei_2022_EmergentAbilitiesLargea,schaeffer_2023_AreEmergentAbilities,liu_2025_SuperpositionYieldsRobust,wu_2024_UshapedInvertedUScalinga,michaelov_2025_LanguageModelBehaviorala,cai_2025_LatentVariableFramework}. Finally, we include research analyzing model internal states, including the analysis of individual `neurons', representation `probing', sub-network and circuit identification, and other related methods \citep[for review, see][]{saphra_2024_Mechanistic}.

\section{The Versioning Problem}
\label{ssec:versioning}

In order to draw inferences about a given model $M$, there is an implicit assumption that $M$ is stable such that past outputs from $M$ are informative about an identically labeled $M$. However, such model stability cannot be taken for granted: \textbf{incomplete versioning in \OURminmodels{} threatens inferences across model executions}. 

ChatGPT, for example, undergoes regular changes that can cause substantial changes to its behavior. These are not always rolled out simultaneously, and at least historically, are not always announced \citep{openai_2025_SycophancyGPT4oWhat,openai_2025_ExpandingWhatWe,openai_2025_ChatGPTReleaseNotes}. Even models that are ostensibly the same are often updated; for example, at the time of writing, there are three versions of the \texttt{GPT-4o} model (\texttt{2024-11-20}, \texttt{2024-08-06}, and \texttt{2024-05-13}), with the default potentially changing with each version release. 

Crucially, one cannot assume that a new version of a model will continue to behave in a substantially similar way to a previous version. Even when the changes involve only the addition of new pretraining data from similar sources, a model's internal state can abruptly change in ways that cannot be anticipated \citep[e.g.,][]{olsson_2022_IncontextLearningInductiona,chen_2024_SuddenDropsLossa}, as can its performance \citep{wei_2022_EmergentAbilitiesLargea,michaelov_2023_EmergentInabilitiesInversea,mckenzie_2023_InverseScalingWhena}. As a concrete example, performance on a benchmark testing a model's ability to determine whether a number is prime or not varies by up to 60\% between the March and June 2023 versions of GPT-3.5 and GPT-4 \citep{chen_2024_HowChatGPTsBehavior,narayanan_2023_GPT4GettingWorse}, and large differences continue to be reported for different versions of the `same' model (\citealp{hackenburg_2025_LeversPoliticalPersuasion}).

There are also other ways in which model behavior is updated over time. For example, some \OURminmodels{} allow user customization based on past interactions \citep{openai_2024_MemoryNewControls,openai_2025_MemoryFAQ,anthropic_2025_ManageClaudesMemory,anthropic_2025_IntroducingClaude4,citron_2025_ReferenceChatsMore}. Moreover, as we discuss at length in \S\ref{ssec:credit_assignment}, additional non-language-model components of `chat' or API pipelines may be updated, potentially altering behavior in unknown and unforeseeable ways. Thus, any execution of a labeled model $M$ may not be fully comparable to any other execution of $M$.

Even with full stability---i.e., if a given $M$ always refers to a specific stable system---the unavailability of previous versions presents a substantial risk to research. \OURminmodels{} are hosted as commercial products rather than scientific artifacts. As such, older models are often deprecated, becoming unavailable to researchers. For example, both versions of GPT-3.5 analyzed by \citet{chen_2024_HowChatGPTsBehavior} are no longer available. Thus, all research on these models, the original GPT-3 models, or any other now-retired model (see, e.g., \citealp{kalyan_2024_SurveyGPT3Family,chang_2024_LanguageModelBehaviora}) expires along with these models, and can no longer be replicated.

Finally, even if the aforementioned stability conditions are met, and the model remains available, specific implementation details, including quantization, hardware details, and software versions can also have an impact on model behavior and are a substantial source of irreproducibility \citep{gundersen_2023_SourcesIrreproducibilityMachine}. While this is an issue for all artificial intelligence systems, a crucial limitation for \OURminmodels{} is that such details are rarely, if ever, disclosed.

\paragraph{When is this a problem?}

The versioning problem applies to all cases of \OUReval{}, \OURcomp{}, and \OURinterp. \textbf{Any findings only reliably apply to the specific executions of the specific instances of each model tested}.

\paragraph{What can be done with closed models?}

The impact of the versioning problem could be mitigated by a closed model provider by providing a user with the option to specify the exact version to be used, committing to persistent access to prior versions (for some reasonable period that enables most comparisons in practice), and supplying assurances of the trustworthiness of these conditions. To our knowledge, none provide these features and assurances. For paper authors, reporting the approximate version and execution date can at least help determine when one should \textit{definitely not} assume the same behavior as reported in a paper.

\paragraph{What can be done with open weights?}

With an \OURopenweight{} and sufficient computing resources, a user can run a known version of that language model. This known version is \textit{fixed} by definition, and thus supports reliable \OUReval{}, \OURcomp{}, and \OURinterp{}. 

\paragraph{Moving beyond open weights alone}
\OURopenweights{} enable stable versioning, but do not guarantee it---it is still possible that an \OURopenweight{} could be updated or become no longer accessible. Thus, it is still valuable to note the date on which such a model was downloaded and from where; and if resources are available, to store a copy of the model that could be redistributed (if the license allows) if needed. Additionally, given the aforementioned instability across software and hardware setups, researchers should report these details and provide the code used to run any experiment where possible.

\section{The Credit Assignment Problem }
\label{ssec:credit_assignment}

A second fundamental barrier to making valid scientific inferences about \OURminmodels{} is uncertainty about their design and implementation. Specifically, \textbf{the behavior of any given \OURminmodel{} cannot reliably be attributed to the language model it contains}.

Language models as standardly defined---systems that compute a probability distribution over possible next words \citep{jurafsky_2025_SpeechLanguageProcessing}---are often only part of the underlying \OURminmodel{} system accessed through chat interfaces and APIs. In addition to the core language model (or models; \citealp{openai_2025_GPT5SystemCard}), the overall system may include system prompts, filtering, guardrails, web search, user customization, knowledge bases, additional APIs, hidden `thinking' or `reasoning' processes, and potentially other features, depending on the product. 

As more components are added, it is unclear whether the system as a whole should be considered a ``language model'' at all. In these cases, the core language model might be better considered a single component of a broader ``compound system'' \citep{zaharia_2024_ShiftModelsCompound} or ``LLM-equipped software tool'' \citep{trott_2024_LanguageModelsVs}. The behavior of such systems can differ from that of an isolated language model---for example, \citet{schick_2023_ToolformerLanguageModels} reports that a system equipped with a calculator API exhibits improved performance on arithmetic tasks over the core language model. Such architectures can be intriguing objects of study in themselves when the components and the mechanisms by which they interact are made transparent by the system designer \citep[e.g.,][]{sumers_2023_CognitiveArchitecturesLanguage}. However, for \OURminmodels{}, it is not possible to separate the role of the core language model from the whole system, threatening inferences about the performance of the language model, its design, and its implementation.

In addition to more explicit compound system components such as internal APIs, there are also more subtle examples. One of these is decoding strategy. The core language model generates a probability distribution over possible next tokens. However, simple ``greedy sampling'' (i.e., incrementally generating the highest-probability token) is rarely used today even with \OURopenweights{}, where temperature sampling, beam search, and nucleus sampling are often the default (see e.g., \citealp{huggingface_2025_Generation,vonplaten_2020_HowGenerateText,labonne_2024_DecodingStrategiesLarge}). Closed model APIs and interfaces do not generally state the decoding strategy used, and given that this is an active research area \citep[e.g.,][]{klein_2017_OpenNMTOpenSourceToolkit,fan_2018_HierarchicalNeuralStory,paulus_2018_DeepReinforcedModel,bachmann_2025_JudgeDecodingFaster,lipkin_2025_FastControlledGeneration}, even the assumption that a publicly-known decoding strategy is being used may be unwarranted. The selection of decoding strategy substantially affects text generation, which threatens the reliability of any assessment of the model component in a \OURminmodel{}. \OURminmodels{} with API access may offer some control over decoding parameters, which may provide some additional insight into core model performance; however, even in these cases, fully deterministic generation is not available \citep{openai_2025_APIReference,google_2025_GenerateContentGemini,anthropic_2025_Messages}.

The credit assignment problem also arises through the injection of hidden system prompts. To the best of our knowledge, the overwhelming majority of \OURminmodels{} currently available through chat interfaces use a hidden and undocumented system prompt, \textit{i.e.}, a specific prompt created by the provider that is prepended to the initial user query. Currently, Claude is one of the few \OURminmodels{} for which any part of the system prompt is provided \citep{anthropic_2025_SystemPrompts}. Crucially, language models have consistently been found to be extremely sensitive to even small differences in their prompts \citep[see, e.g.,][]{sclar_2024_QuantifyingLanguageModels,shi_2023_LargeLanguageModels,zhuo_2024_ProSAAssessingUnderstanding,polo_2024_EfficientMultipromptEvaluation,razavi_2025_BenchmarkingPromptSensitivity,errica_2025_WhatDidWronga,chatterjee_2024_POSIXPromptSensitivity,zhan_2024_UnveilingLexicalSensitivity}. Thus, unknown variations in system behavior due to system prompts confound valid comparisons between models. 

Finally, post-processing strategies, such as filters that provide ``guardrails'' to protect against undesirable outputs, are common in \OURminmodels{}. These are generally undocumented, and can significantly alter the output in ways that cannot be anticipated by the user \citep[see, e.g.,][]{bourg_2024_GenerativeAITrustworthy}, and further limit successful credit assignment.

\paragraph{When is this a problem?} Carrying out \OUReval{}, \OURcomp{}, or \OURinterp{} with a language model embedded in a compound system requires being able to disentangle the behavior of the language model from the behavior of the compound system. As discussed, this is not possible with \OURminmodels{}, and thus all of these are threatened.

\paragraph{What can be done with closed models?} There is currently no reliable way to mitigate the credit assignment problem for language models embedded in \OURminmodel{} compound systems. In principle, model providers of \OURminmodels{} could host either language models that are \textit{not} embedded in compound systems, as in the case of the original GPT-3 \citep{brown_2020_LanguageModelsArea}, or allow more complete customization of the compound system in which the models are embedded.

Alternatively, if the versioning problem were resolved, it could be possible to carry out \OUReval{} and \OURcomp{} on \OURminmodels{} as complete commercial products, but without the ability to make any reliable inferences about \textit{why} a given system behaves in a certain way or performs better than another.

\paragraph{What can be done with open weights?} 
Having an \OURopenweight{} is sufficient to run a specified language model separately from the compound system in which it is embedded, allowing reliable \OUReval{} and \OURcomp{}. Thus, a user with resources to run models independently can resolve the ambiguous credit assignment problem for a given language model.

\paragraph{Moving beyond open weights alone}

\OURopenweights{} allow researchers to characterize the behavior of a language model independently of the compound system in which it is embedded. Taking this a step further, in a fully specified compound system including the \OURopenweight{} and fully open versions of all other components, the credit assignment problem can be resolved for the other components as well.

\section{The Information Restriction Problem}
\label{ssec:output_only}

The final limitation we discuss in making robust scientific inferences with fully \OURminmodels{} is that in many cases, \textbf{\OURminmodels{} simply do not provide access to information necessary to reliably and efficiently achieve researchers' inferential goals}.

In some ways, this is true by definition. \OURminmodels{} only provide model output, and thus, only \OURcomp{} and \OURinterp{} research that can be carried out on the basis of \OUReval{} is possible even as a starting point. However, there are more subtle impacts of using a \OURminmodel{} in research, specifically in terms of how it restricts the kinds of \OUReval{} questions that can be studied. Since a \OURminmodel{} provides only text output, there is only one way to evaluate it: provide it with an input, allow it to generate an output, and then use some procedure to evaluate this output. In principle, assuming the Versioning and Credit Assignment problems were resolved, this could be sufficient for making reliable \OUReval{} inferences, but the scope of these is limited.

Consider a practical use case such as medical diagnosis (for concerns about such a use of language models, see, e.g., \citealp{raji_2022_FallacyAIFunctionalitya,ullah_2024_ChallengesBarriersUsing,shool_2025_SystematicReviewLarge}). For such a critical use case, it would be crucial to have estimates of confidence in the top prediction. However, the vast majority of approaches for estimating this rely on examining the probability language models assign their outputs (see \citealp{hills_2023_UsingLogprobs,jiang_2021_HowCanWe,chen_2023_CloseLookCalibration,geng_2024_SurveyConfidenceEstimation,liu_2025_UncertaintyQuantificationConfidence}), which is not available for fully \OURminmodels{}. And while one could prompt a language model to provide not only the most likely diagnosis but also the probability of the second and third most likely (and so on), there is no guaranteed relationship between the text generated in response to such a prompt and the probability assigned to each of these in the output distribution---language models cannot directly access the probabilities of their own output probability distribution.

\OURminmodels{} are also generally not amenable to diagnostic benchmarks assessing their general `behavior' or `capacities'. For example, if one wants to investigate whether a language model is able to generate grammatical text, the most reliable way to do so is to compare the probability a language model assigns to a grammatical text sequence to that assigned to a minimally-edited ungrammatical sequence \cite{hu_2023_PromptingNotSubstitutea}. Adapting this to a format where a model is presented with both sentences and asked to judge which is grammatical changes the nature of the task to a meta-linguistic one that relies on being able to answer questions, which previous work suggests advantages larger and more fine-tuned models \citep{hu_2023_PromptingNotSubstitutea,hu_2024_AuxiliaryTaskDemands,chung_2024_ScalingInstructionFinetunedLanguage,song_2025_LanguageModelsFail}, which could lead to over-estimating the performance of (often larger and more heavily fine-tuned) \OURminmodels{} relative to most \OURopenweights{}. Similarly, when assessing whether a model is able to generate text that is consistent with the physical world, it is more informative to know whether, following a text extract such as \textit{Marissa forgot to bring her pillow on her camping trip. As a substitute for her pillow, she filled up an old sweater with}, a model is able to assign a higher probability to the unlikely but physically possible \textit{leaves} than the impossible \textit{water}, than to know the model's top prediction \citep{glenberg_2000_SymbolGroundingMeaninga,jones_2022_DistrubutionalSemanticsStilla}, something which is again impossible to do based only on text output. These are only two examples, but a wide range of \OUReval{} approaches---including many current benchmarks (see, e.g., \citealp{srivastava_2023_ImitationGameQuantifyinga,liang_2023_HolisticEvaluationLanguage,biderman_2024_LessonsTrenchesReproducibleb})---require access to more than just the top prediction (or a limited set of top predictions) for fine-grained assessments of model reliability and robustness. These are all by definition impossible for \OURminmodels{}.

It is also important to not understate the practical limitations of evaluation based on text outputs: outputs are both high-dimensional (a sequence of one or more tokens in an often very large vocabulary) and the flexibility of language (or, e.g., code) means that there are often a vast number of sequences of tokens that are correct or even roughly equivalent. All of the automatic methods that have been proposed for assigning labels or scores to output text sequences---including rule-based methods (e.g., exact string matching; see \citealp{biderman_2024_LessonsTrenchesReproducibleb}), heuristic-based methods (e.g., BLEU; \citealp{papineni_2002_BleuMethodAutomatic}), classifier-based methods (e.g., BERTScore; \citealp{zhang_2020_BERTScoreEvaluatingText}), and prompting a second model to rate the output (i.e., LLM-as-judge; \citealp{lin_2022_TruthfulQAMeasuringHowa})---have been found to be brittle, noisy, and to exhibit a range of systematic issues (see, e.g., \citealp{murray_2018_CorrectingLengthBias,hanna_2021_FineGrainedAnalysisBERTScore,bavaresco_2025_LLMsInsteadHuman,baumann_2025_LargeLanguageModel}). Even human-based evaluation can be problematic given known biases (see, e.g., \citealp{chen_2024_HumansLLMsJudgea}) and often may not be feasible due to the expertise needed to rate increasingly difficult tasks (see, e.g., \citealp{rein_2024_GPQAGraduateLevelGoogleProof}). Thus, \OUReval{} is already substantially limited by the scoring method confound.

\paragraph{When is this a problem?}
As discussed, text output alone does not generally provide the information necessary for \OUReval{} and \OURcomp{} inferences. When a researcher aims to go beyond simple evaluation approaches such as calculating accuracy of the top prediction, outputs of \OURminmodels{} will thus not be sufficient. Furthermore, \OURminmodels{} do not provide internal states, and thus, \OURinterp{} is also limited---approaches that rely on internal states are impossible to implement, and those that rely on model output suffer from the same issues as \OUReval{}.

\paragraph{What can be done with closed models?} 
As discussed, limited and noisy model evaluations can be performed using the text sequences provided by \OURminmodels{}. Output sequence probabilities (for directly calculating model confidence) are provided for some API-based models, though this is far from universal, and as discussed in \S\ref{ssec:credit_assignment}, calculation methods are not transparent or deterministic (see \citealp{openai_2025_APIReference,dong_2025_UnlockGeminisReasoning,wise_2024_FeatureAddSupport,anthropic_2025_TokenCounting}). We are not aware of any \OURminmodel{} for which specific (user-selected) token probabilities or full output probability distributions are provided since the now-deprecated GPT-3 \citep{brown_2020_LanguageModelsArea,ouyang_2022_TrainingLanguageModelsb} and \textsc{Jurassic-1} \citep{lieber_2021_Jurassic1TechnicalDetails}, nor any model for which the internal states are provided. The former would enable researchers to calculate and compare the probabilities of predefined outputs, allowing for the \OUReval{} of model behavior outside the highest-probability outputs and thus enabling \OURcomp{} and limited \OURinterp{} research. 

\paragraph{What can be done with open weights?}
\OURopenweights{} provide full access to the model output distribution and internal states, which is sufficient information for \OUReval{} and \OURcomp{}, and for many, but not all, \OURinterp{} research questions.

\paragraph{Moving beyond open weights alone}
\OURinterp{} often requires full access to model outputs and internal states, but research questions about the role of training data on model behavior often also require access to some combination of training data, checkpoints, and gradients \citep[e.g.,][]{biderman_2023_PythiaSuiteAnalyzinga,groeneveld_2024_OLMoAcceleratingSciencea,lesci_2024_CausalEstimationMemorisation,liu_2025_OLMoTraceTracingLanguage,chang_2025_ScalableInfluenceFact}. Carrying out an analysis on multiple training runs of the same model with different random seeds can also facilitate stronger inferences \citep{sellam_2021_MultiBERTsBERTReproductions,vanderwal_2024_PolyPythiasStabilityOutliersa}.

\section{Discussion, Limitations, and Possible Objections}
\label{sec:discussion}

Our analysis reveals that in their current state, \OURminmodels{} are generally unsuitable for \OUReval{}, \OURcomp{}, and \OURinterp{} research. Below, we address possible concerns and highlight cases were the use of \OURminmodels{} may be preferred.

\subsection{What if only \OURminmodels{} show state-of-the-art capabilities?}
As discussed in \S\ref{ssec:versioning}--\S\ref{ssec:output_only}, \OURminmodels{} do not generally provide sufficient information to enable reliable \OURcomp{}. Thus, any claim about the performance of a \OURminmodel{} with respect to a given task or set of tasks requires convincing evidence that threats to robustness were mitigated. Without such evidence, these claims should not be taken at face value. This has an important further implication: \textbf{It is generally \textit{not} appropriate to choose to use a given \OURminmodel{} rather than an \OURopenweight{} on the basis of reported performance alone}.

\subsection{What about barriers to entry in the use of \OURopenweights{}?}

As products, \OURminmodels{} are designed to be easy to use. Many current model providers also have free or lower-cost \OURminmodels{} available for use. Meanwhile, locally running an \OURopenweight{} can require substantial computational resources and expertise. These factors can each pose barriers to researchers conducting research on language models. 

However, these issues do not invalidate the concerns raised in \S\ref{ssec:versioning}--\S\ref{ssec:output_only}. If none of the inferences drawn from such work can be considered reliable, then it is not clear what is learned by carrying it out. More constructively, it is worth noting that there are (partial) solutions to both the resource and expertise concerns. One is the development of software frameworks that enable more streamlined use of contemporary language models---most notably the \textit{transformers} \citep{wolf_2020_TransformersStateoftheArtNaturala} package, which allows the same pipelines to be used to generate text, calculate output probabilities, analyze model states, and train language models of a range of architectures.
Model hosting and inference has also decreased in price. For example, companies and organizations allow users to cheaply generate outputs with currently-existing \OURopenweights{}.\footnote{For a regularly-updated list, see \url{https://huggingface.co/docs/inference-providers/en/index}} These all resolve the versioning problem (see \S\ref{ssec:versioning}), and some (such as \textit{Hugging Face Inference}\footnote{\url{https://huggingface.co/learn/cookbook/en/enterprise_dedicated_endpoints}}) provide log-probabilities and control over generation such that many of the possible credit assignment and information restriction issues may be avoided. Resources such as NSF National Deep Inference Fabric (NDIF) in the USA \citep{fiotto-kaufman_2024_NNsightNDIFDemocratizing} and the European Deep Inference Fabric (eDIF) in Europe \citep{guggenberger_2025_EDIFEuropeanDeep} also provide researchers with the ability to freely access internal states of widely-used \OURopenweights{}, and tools such as \textit{NNsight} \citep{fiotto-kaufman_2024_NNsightNDIFDemocratizing} provide streamlined methods for manipulating them.

\subsection{When Should \OURminmodels{} be Considered for Scientific Research?}
\label{ssec:closed_acceptable}

The main conclusion of our analysis in \S\ref{ssec:versioning}--\S\ref{ssec:output_only} is that \OURminmodels{}---at least in their format---are not suitable for most cases of \OUReval{}, \OURcomp{}, and \OURinterp{}. However, there are also cases where it can be more appropriate to use \OURminmodels{} instead of \OURopenweights{}. We present some examples below.

\paragraph{Closed models for one-off solution generation and existence proofs} The apparent capabilities of \OURminmodels{} are often impressive. As discussed above, \OURminmodel{} performance on a given task cannot predict future performance, and cannot generally be fairly compared to that of other models. Nonetheless, a high performance obtained from a \OURminmodel{} can be used as an existence proof that a system can perform a given task at a certain level; and thus can serve as an upper baseline (however unstable) when comparing more open models (e.g., \citealp{moslem_2023_AdaptiveMachineTranslation,zhu_2024_MultilingualMachineTranslation,tang_2025_LargeLanguageModels,xu_2025_ChatQA2Bridging}). This is analogous to using a task-specific system, such as Google Translate, as an upper baseline when evaluating systems' machine translation capabilities \citep[e.g.,][]{nguyen_2022_KC4MTHighQualityCorpus,moslem_2023_AdaptiveMachineTranslation,garcia_2023_UnreasonableEffectivenessFewshot,zhu_2024_MultilingualMachineTranslation}. Similarly, it is reasonable to use \OURminmodels{} to generate one-off verifiable solutions, as in the case of human-computer collaboration in the creation of mathematical proofs \citep[e.g.,][]{schmitt_2025_ExtremalDescendantIntegrals,feng_2026_SemiAutonomousMathematicsDiscovery}.

\paragraph{Closed models as objects of study in socio-technical research.} Today, hundreds of millions of individuals interact with language models in various ways. Thus, from the perspective of ecological validity, researchers interested in the effects of AI systems on individuals and society should generally focus on currently-deployed systems, though this should be explicitly justified. In the past, researchers have investigated how humans interact with or perceive text generated by systems such as Google Translate \citep{kol_2018_GoogleTranslateAcademic}. Similar research on \OURminmodels{} \citep{stiennon_2020_LearningSummarizeHumana,jannai_2023_HumanNotGamifieda,costello_2024_DurablyReducingConspiracy,jones_2024_DoesGPT4Passa,jones_2025_LargeLanguageModels,rathi_2025_GPT4JudgedMore} has also yielded valuable insights.

It is also appropriate to look at currently used systems when studying possible harms caused by model outputs, especially since even an existence proof of a specific model behavior pattern can have immediate real-world implications. Again, there are parallels between research on more traditional software tools such as Google Translate (e.g., gender bias; \citealp{prates_2020_AssessingGenderBias}) and on contemporary \OURminmodels{} \citep{kotek_2023_GenderBiasStereotypes,ghosh_2023_ChatGPTPerpetuatesGender,salinas_2023_UnequalOpportunitiesLarge,lee_2024_LargeLanguageModels,ding_2025_GenderBiasLarge}. The conversational nature of most contemporary \OURminmodels{} also potentially enables new types of harm. For example, such systems have been shown to be `deceptive'---they can produce text that is either inconsistent with the truth, misleading, or does not accurately reflect the model's past or future behavior \citep{hagendorff_2024_DeceptionAbilitiesEmerged,jones_2024_LiesDamnedLies,dogra_2025_LanguageModelsCan,taylor_2025_LargeLanguageModels}. Such work has limited generalizability but still serves as an existence proof establishing a potentially harmful behavior or the ability to elicit a given effect in a human participant.

\subsection{Do \OURopenweights{} guarantee reliable scientific inference?}
As we note throughout the paper, \OURopenweights{} inherently avoid many inferential limitations of \OURminmodels{}. Nonetheless, in some cases, additional types of information may be needed. Perhaps most prominent of these is training data. If there is an overlap in the data used to train and test the model (including if the model is trained on user data that includes test set samples; see \citealp{balloccu_2024_LeakCheatRepeat}),
this can drastically impact model behavior. If the overlap is unknown, it can present a substantial confound \citep{brown_2020_LanguageModelsArea,blevins_2022_LanguageContaminationHelps,roberts_2023_CutoffLongitudinalPerspective,mehrbakhsh_2024_ConfoundersInstanceVariation,dong_2024_GeneralizationMemorizationData,yao_2024_DataContaminationCan,jiang_2024_DoesDataContamination}. Where possible, researchers should select models that make available complete information characterizing model design, implementation, and execution. Today, the most `open' models extant are \OURopenweights{} that also make available open data, training code, and training checkpoints, such as Pythia \citep{biderman_2023_PythiaSuiteAnalyzinga,vanderwal_2024_PolyPythiasStabilityOutliersa}, OLMo \citep{groeneveld_2024_OLMoAcceleratingSciencea,olmoteam_2025_2OLMo2}, and SmolLM \citep{benallal_2024_SmolLMBlazinglyFast,benallal_2025_SmolLM2WhenSmola}.

Open models are the foundation for reliable, comparable, and robust inferences. 
However, their use alone cannot guarantee trustworthy inferences. For example, the use of an open model cannot guarantee that test data is relevant and selected appropriately; that the model performs well on the inputs; that features produced from the resulting outputs are valid measures of relevant constructs and concepts; or that discovered associations between measurements can be reliably causally interpreted. Reliable inference requires thoughtful research design and systematic implementation. Open models merely provide the opportunity to interpret results transparently. 

\section{Recommendations} 
\label{sec:recommendations}
In this paper, we have demonstrated how the choice of whether to used a \OURminmodel{} or \OURopenweight{} is a major methodological choice that has substantial impact on how the results of a study should be interpreted. Based on this, we provide the recommendations below, with further guidance provided in the sample checklist included in Appendix \ref{app:checklist}.

As a \textbf{researcher}, if it is possible to use an \OURopenweight{}, do. If it is not, systematically review the threats of using a \OURminmodel{} to inferential validity, consider the potential impact on the core questions you aim to answer, and select appropriate mitigation. Document this analysis and mitigation, along with your conclusions, for users of your research. 

As a \textbf{reader}, be skeptical of the results from or comparisons across closed models for scientific inferences, unless a closed model is necessary for the object of study and relevant threats to inference are mitigated.

As a \textbf{reviewer}, consider information characterizing the extent to which any models used have stable versioning and possible credit assignment as necessary parts of methods section. Require that inferences reported using closed models be justified by appropriate mitigations, and ensure that any conclusions made are appropriately scoped given these issues. Failure to do so should be considered valid grounds for rejection. As an \textbf{editor} or \textbf{program chair}, consider requiring authors to fill out a checklist (see Appendix \ref{app:checklist}) covering these issues as part of the submission process.

\section{Conclusions}
The amount of information available about a language model is crucial for determining the kinds of inferences that can be drawn when using it in scientific research. However, despite it being a significant methodological decision, the choice about whether to use a closed or more open model---and the question of how  to interpret the results based on this---is often left unexamined and undocumented. We provide a framework for considering when it is appropriate to use closed models and when more open models are needed. We summarize the key results of our analysis in \Cref{fig:summary_table}, and provide a sample checklist in Appendix \ref{app:checklist} to aid in implementing our framework in research.

\clearpage

\bibliography{references}
\bibliographystyle{apalike}

\clearpage

\appendix
\section{Proposed Pretrained Language Model Checklists}
\label{app:checklist}
\begin{enumerate}
\item \textbf{Generalizable Claims about Language Models and Related Systems}
    \item[] Question: Do any of the main results or claims in the paper state, imply, or depend on the generalizability of the `behavior' (including outputs, activations, and other mechanisms) of one or more pretrained models (in the sense of models trained by anyone other than the authors)? If the answer to this question is \answerNo{}, questions 2--3 should be answered with \answerNA{}.
    \item[] Answer: \justificationTODO{}
    \item[] Justification: \justificationTODO{}
    \item[] Guidance:
    \begin{itemize}
        \item The answer \answerNo{} means that the paper no generalizable claims are made about language models, for example:
        \begin{itemize}
            \item The paper does not refer to language models. 
            \item The paper uses language models or language-model-based systems only to generate candidate solutions that are then evaluated by other means, and makes no general claims about model performance or behavior.
            \item The paper claims only an existence proof that at least one language model or language-model-based system behaved in a specific way at one specific time. Any claim as to the generalizability of this (e.g., that this behavior could be expected to occur at a future point in time) should lead to a response of \answerYes{}. This includes implicit claims of generalizability such as attributing specific capacities or traits to a model or set of models.
        \end{itemize}
    \end{itemize}

\item \textbf{Language Models as Stable Artifacts}
    \item[] Question: For any of the systems about which generalizable claims are made (see Q1), is the system clearly identified, known to be stable, and available for external validation?
    \item[] Answer: \justificationTODO{}
    \item[] Justification: \justificationTODO{}
    \item[] Guidance:
    \begin{itemize}
        \item The answer \answerNA{} means that that no generalizable claims are made about any language model or language-model-based system.
        \item The answer \answerYes{} means that sufficient information provided for the reader to use an identical version of the system described in the paper. Note that this generally does not include any model accessible only through a chat interface or API, unless there is a specific guarantee that the complete system (including preprocessing, postprocessing, decoding algorithm, guardrails, etc.) is fully stable. For open-weight systems, this should consist at the very least of a link to where the model and any other components of the system (if applicable) are available and an access or download date.
        \item The answer \answerNo{} covers all other cases, and requires a strong justification that should be discussed in the paper. For example:
        \begin{itemize}
            \item The paper focuses on analyzing human interactions with language models, it is important to study interactions between humans and systems that are actually used in practice, and the system in question is among the most prevalent.
            \item The paper focuses on studying larger sociotechnical systems including humans and computational systems, the combination of which is not possible replicate exactly.
        \end{itemize}
    \end{itemize}

\item \textbf{Sufficient Model Information}
    \item[] Question: For any of the systems about which generalizable claims are made (see Q1), is sufficient model information provided to support or otherwise justify the claim?
    \item[] Answer: \justificationTODO{}
    \item[] Justification: \justificationTODO{}
    \item[] Guidance:
    \begin{itemize}
        \item The answer \answerNA{} means that that no generalizable claims are made about any language model or language-model-based system.
        \item The justification should indicate the location where the approach taken is justified and explained. We encourage reviewers carefully to consider whether these justifications are sufficient. Below we provide guidance about what kind of information is generally needed in order to make a claim regarding one of the following kinds of inferential goal:
            \begin{enumerate}
                \item \textbf{Model Evaluation--Benchmark Score}: The only aim is to calculate the score of a model on one or more specific datasets in order to compare it to a baseline.
                \begin{itemize}
                    \item Required information: Generally, whatever information is required to score the model is sufficient. Thus, in principle, text output alone could be sufficient. However, in many cases, evaluations rely on comparing the probability a model assigns to different text strings, which requires access to the model's output distribution; and adapting such a task to a text generation task can change what the task measures.
                \end{itemize}
                \item \textbf{Model Evaluation--Real-World Use}: The goal is to evaluate the performance of the model on a real-world task.
                \begin{itemize}
                    \item Required information: In some cases, whatever information is required to calculate a score for the model (e.g., text output) is sufficient, but generally, fine-grained analyses, including analyses of the errors made, how confident the model was, how close the model was to the correct response, or robustness checks are needed, which requires information about the output distribution.
                \end{itemize}
                \item \textbf{Model Evaluation--Behavior/Capability Assessment}: The goal is to make a generalizable claim about model behavior, for example whether a model is generally capable of a specific type of task, exhibits a systematic bias, or is robust to unlikely inputs.
                \begin{itemize}
                    \item Required information: In the case of very careful analyses of model outputs with a strong theoretical basis, model output alone may be sufficient, but in the majority of cases, the output distribution is needed in order to carry out the kinds of analyses described in (b), and additional model design information (e.g., training data) may also be required.
                \end{itemize}
                \item \textbf{Model Comparison--Model Selection}: The goal is to determine which model performs best on a given dataset.
                \begin{itemize}
                    \item Required information: In principle, text output, but often also the output probability distribution (see (a)).
                \end{itemize}
                \item \textbf{Model Comparison--Other}: The goal is to compare model performance or behavior in some way beyond simple benchmark performance, for example, on a real-world task, or in terms of their robustness or general capability at a specific type of task.
                \begin{itemize}
                    \item Required information: In general, any form of comparison beyond ranking models on a given task will require access to the model's output distribution (see (b)). Additionally, any form of comparison that treats language models as something other than text generators will require information about model design (e.g., it is generally inappropriate to compare a closed language-model-based product that can search the internet to a language model in isolation).
                \end{itemize}
                \item \textbf{Model Interpretability--Model Design}: The goal is to determine whether a specific design feature or set of design features (architecture, training details, etc.) has an impact on model behavior.
                \begin{itemize}
                    \item Required information: Requires sufficient information about model design (including, e.g., training data) to determine that there isn't a confound. Fine-grained analysis of behavior may also require access to the model's output distribution.
                \end{itemize}
                \item \textbf{Model Interpretability--Explanation}: The goal is to understand why a model behaves in a specific way given a specific input.
                \begin{itemize}
                    \item Required information: Requires sufficient information about model design (including, e.g., training data) to determine that there isn't a confound. Fine-grained analysis of behavior may also require access to the model's output distribution.
                \end{itemize}
            \end{enumerate}
    \end{itemize}

\end{enumerate}


\end{document}